\documentclass{llncs}
\pdfoutput=1
\usepackage{amsmath, amsfonts, graphicx, booktabs, tabularx, multirow, siunitx, xcolor, soul, bm}
\usepackage{hyperref}
\graphicspath{{../figures/}}

\begin{document}
\title{Chest X-ray Report Generation through Fine-Grained Label Learning}
\author{Tanveer Syeda-Mahmood, Ken C. L. Wong, Yaniv Gur, Joy T. Wu, \\ Ashutosh Jadhav, Satyananda Kashyap, Alexandros Karargyris,  Anup Pillai, Arjun Sharma, Ali Bin Syed, Orest Boyko, Mehdi Moradi\\ {\tt\small stf@us.ibm.com}
}


\institute{IBM Almaden Research Center, San Jose, CA} 

\maketitle

\begin{abstract}
Obtaining automated preliminary read reports for common exams such as chest X-rays will expedite clinical workflows and improve operational efficiencies in hospitals.  However, the quality of reports generated by current automated approaches is not yet clinically acceptable as they cannot ensure the correct detection of a broad spectrum of radiographic findings nor describe them accurately in terms of laterality, anatomical location, severity, etc.  In this work, we present a domain-aware automatic chest X-ray radiology report generation algorithm that learns fine-grained description of findings from images and uses their pattern of occurrences to retrieve and customize similar reports from a large report database.  We also develop an automatic labeling algorithm for assigning such descriptors to images and build a novel deep learning network that recognizes both coarse and fine-grained descriptions of findings.  The resulting report generation algorithm significantly outperforms the state of the art using established metrics.\end{abstract}

\section{Introduction}
Chest X-rays are the most common imaging modality read by radiologists in hospitals and tele-radiology practices today. With advances in artificial intelligence, there is now the promise of obtaining automated preliminary reads that can expedite clinical workflows, improve accuracy and reduce overall costs. Current automated report generation methods are based on image captioning approaches of computer vision \cite{vinyals2015show,xu2015show} and use an encoder-decoder architecture where a Convolutional Neural Network (CNN) is used to encode images into a set of semantic topics \cite{jing2017automatic} or limited findings \cite{jiebo} and a Recurrent Neural Network (RNN) decoder or a hierarchical LSTM generates the most likely sentence given the topics \cite{harzig2019addressing,rennie2017self,krause2017hierarchical,wang2018tienet,li2018hybrid,jing2017automatic}. Other approaches have leveraged template sentences to aid in paraphrasing and report generation \cite{li2018hybrid,gale2018producing,li2019knowledge}. Recent approaches have also emphasized the role of clinical accuracy measured loosely through clinical correlation between disease states in the objective functions \cite{szolovits}.  

Despite the progress, the quality of reports generated by current approaches is not yet clinically acceptable as they do not ensure the correct detection of a comprehensive set of findings nor the description of their clinical attributes such as laterality, anatomical location, severity, etc. The emphasis is usually more on the report language generation than the visual detection of findings. In this paper we take a new approach in which we train deep learning networks on a large number of detailed finding labels that represent an in-depth and comprehensive characterization of findings in chest X-ray images.  An initial set of core findings label vocabulary were derived through a multi-year chest X-ray lexicon building effort involving several radiologists and clinical experts.  The detailed finding labels were then automatically derived from their associated radiology reports through a concept detection and phrasal grouping algorithm that associates detailed characterization modifiers with the initially identified core findings using natural language analysis. The resulting labels with large image support were used to train a novel deep learning network based on feature pyramids. Given a new chest X-ray image, the joint occurrence of detailed finding labels is predicted as a pattern vector from the learned model and is matched against a pre-assembled database of label patterns and their associated reports. Finally, the retrieved report is post-processed to remove mentioned findings whose evidence is absent in the predicted label pattern.

\begin{figure}
\begin{center}
   \includegraphics[width=0.8\linewidth,clip]{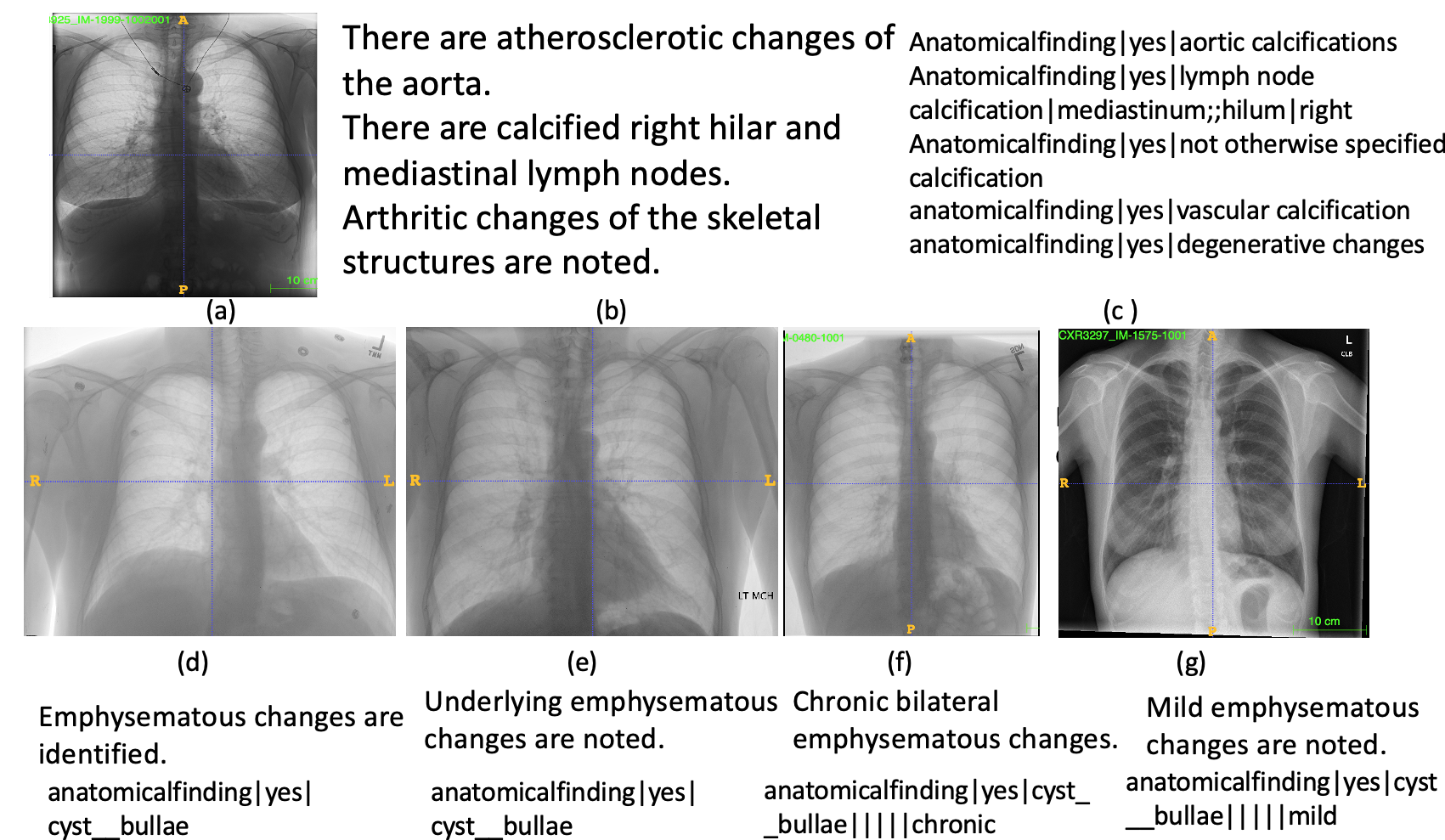}
\end{center}
   \caption{Illustration of the finer description labels for capturing the essence of reports.}
\label{reportexample1}
\end{figure}	

\section{Describing images through fine finding labels (FFL)}
Consider the chest X-ray image shown in Figure~\ref{reportexample1}a. Its associated report is shown in Figure~\ref{reportexample1}b.  In order to automatically produce such sentences from analyzing images, we need image labels that cover not only the core finding, such as opacity, but also its laterality, location, size, severity, appearance, etc.  Specifically, a full description of the finding can be denoted by a fine finding label (FFL) as 
\begin{equation}
F_{i}=<T_{i}|N_{i}|C_{i}|M_{i}^{*}>
\end{equation}
\noindent where $F_{i}$ is the FFL label, $T_{i}$ is the finding type, $N_{i}=yes|no$ indicates a positive or negative finding (i.e is present versus absent), $C_{i}$ is the core finding itself, and $M_{i}$ are one or more of the possible finding modifiers. The finding types in chest X-rays are adequately covered by six major categories namely, anatomical findings, tubes and lines and their placements, external devices, viewpoint-related issues, and implied diseases associated with findings.  The vocabulary for core findings as well as possible modifiers was semi-automatically assembled through a multi-year chest X-ray lexicon development process in which a team of 4 clinicians including 3 radiologists,  iteratively searched through the best practice literature such as Fleishner Society guidelines \cite{hansell2008fleischner} and used every day use terms to expand the vocabulary by examining a large dataset of 220,000 radiology reports in a vocabulary building tool \cite{dla} addressing abbreviations, misspellings, semantical equivalence and ontological relationships.  Currently, the lexicon consists of over 11000 unique terms covering the space of 78 core findings and 9 modifiers and represents the largest set of core findings assembled so far. The set of modifiers associated with each core finding also depends on the finding type and the FFL label syntax captures these for various finding types. 

The FFL labels capture the essence of a report adequately as can be seen in Figure~\ref{reportexample1}c and comparing with the actual report in Figure~\ref{reportexample1}b.  Further, if the FFL labels are similar, a similarity is also implied in the associated reports. Figure~\ref{reportexample1}d-g show examples of similar reports all of which are characterized by similar FFL patterns. Thus if we can infer the FFL labels from the visual appearance of findings in chest X-ray images, we can expect to generate an adequate report directly from the labels.

 \subsection{Extraction of FFL Labels from reports}
The algorithm for extracting FFL labels from sentences in reports consists of 4 steps, namely, (a) core finding and modifier detection,  (b)  phrasal grouping, (c) negation sense detection, (d) pattern completion. 

\begin{figure}
\begin{center}
   \includegraphics[width=0.90\linewidth]{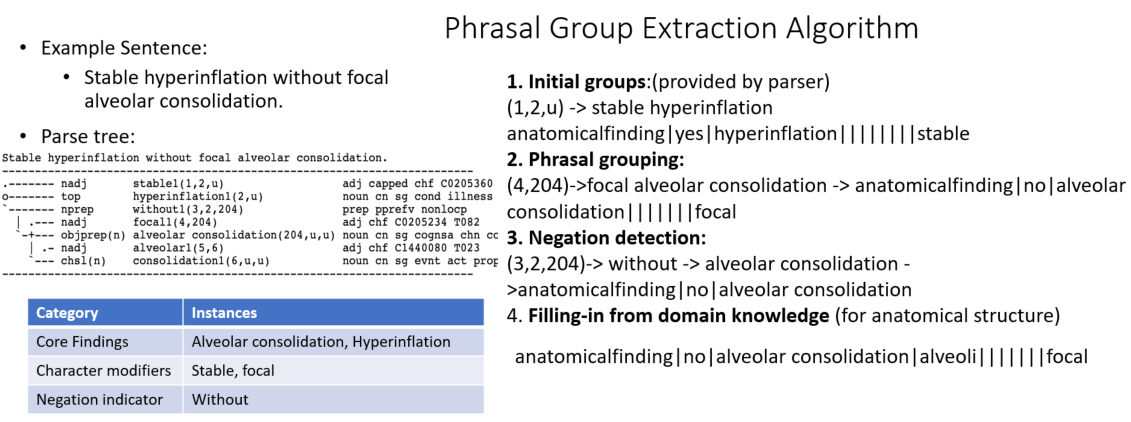}
\end{center}
\caption{Illustration of the dependency parse tree and phrasal grouping.}
   \label{dependency}
\end{figure}

The vocabulary of core findings from lexicon and their synonyms were used to detect core concepts in sentences of reports using the vocabulary-driven concept extraction algorithm described in \cite{guo2017efficient}. To associate modifiers with relevant core findings,  we used a natural language parser called the ESG parser \cite{esg} which performed word tokenization and morpholexical analysis to create a dependency parse tree for the words in a sentence as shown in Figure~\ref{dependency}. The initial grouping of words is supplied directly by the parse tree such as the grouping of terms 'alveolar' and 'consolidation' into one term 'alveolar consolidation' shown in Figure \ref{dependency}. Further phrasal grouping is done by clustering the lemmas using word identifiers specified in the dependency tree. For this, a connected component algorithm is used on the word positions in slots, skipping over unknowns (marked with u in tuples). This allows all modifiers present within a phrasal group containing a core finding to be automatically associated with the finding.  For example, the modifier 'stable' is associated with the core finding 'alveolar consolidation' in Figure~\ref{dependency}. The modifiers in phrasal groups that do not contain a core finding are associated with the adjacent phrasal groups that contain a core finding. 

To determine if a core finding is a positive or negative finding (e.g. "no pneumothorax"), we use a two-step approach that combines language structuring and vocabulary-based negation detection as described in \cite{guo2017efficient}. The negation pattern detection algorithm iteratively identifies words within the scope of negation by iteratively expanding neighborhood of seed negation terms by traversing the dependency parse tree of a sentence. The details are described in \cite{guo2017efficient}. 

The last step completes the FFL pattern using a priori knowledge captured in the lexicon for the associated anatomical locations of findings when these are not specified in the sentence itself as shown in Figure~\ref{dependency} where the term "alveoli" is inserted from the knowledge of the location of the finding 'alveolar consolidation'. Thus the final FFL label produced may show more information than the original sentence from which it was extracted. In addition, the name of the core finding may be ontologically rolled up to the core findings as seen in Figure~\ref{reportexample1} for 'emphysema'.

The FFL label extraction algorithm was applied to all sentences from a collection of 232,964 reports derived from MIMIC-4 \cite{mimic-4} and NIH \cite{wang2017chestx} datasets, to generate all possible FFL patterns from the Findings and Impression sections of reports.  A total of 203,938 sentences were processed resulting in 102,135 FFL labels. By retaining only those labels with at least 100 image support, a total of 457 FFL labels were selected. As shown in the Results section, the label extraction process is highly accurate, so that spot check clinical validation is sufficient for use in image labeling.  Since the FFL labels were seeded by clinically selected core findings, nearly 83\% of all FFL labels extracted could be mapped into their nearest counterpart in the 457 FFL Label set. Thus the set of 457 labels were found sufficient to cover a wide variety in spoken sentences and were used as labels for building deep learning models. Of these, 78 were the original core labels (called the CFL labels) given by clinicians, and the remaining were finer description labels with modifiers extracted automatically. 

\subsection{Learning FFL labels from images}
The learning of FFL labels  from chest X-rays is a fine-grained classification problem for which single networks used for computer vision problems may not yield the best performance, particularly since large training sets are still difficult to obtain. The work in \cite{Conference:Nguyen:ISCAS2018} shows that concatenating different ImageNet-pretrained features from different networks can improve classification on microscopic images. Following this idea, we combine the ImageNet-pretrained features from different models through the Feature Pyramid Network in \cite{Conference:Lin:CVPR2017}. This forms the multi-model feature pyramid which combines the features in multiple scales. The VGGNet (16 layers) \cite{Journal:Simonyan:arXiv2014} and ResNet (50 layers) \cite{Conference:He:CVPR2016} are used as the feature extractors. As nature images and chest X-rays are in different domains, low-level features are used. From the VGGNet, the feature maps with 128, 256, and 512 channels are used, which are concatenated with the feature maps from the ResNet of the same spatial sizes which have 256, 512, and 1024 feature channels.

\begin{figure}[t]
\begin{center}
   \includegraphics[width=0.9\linewidth]{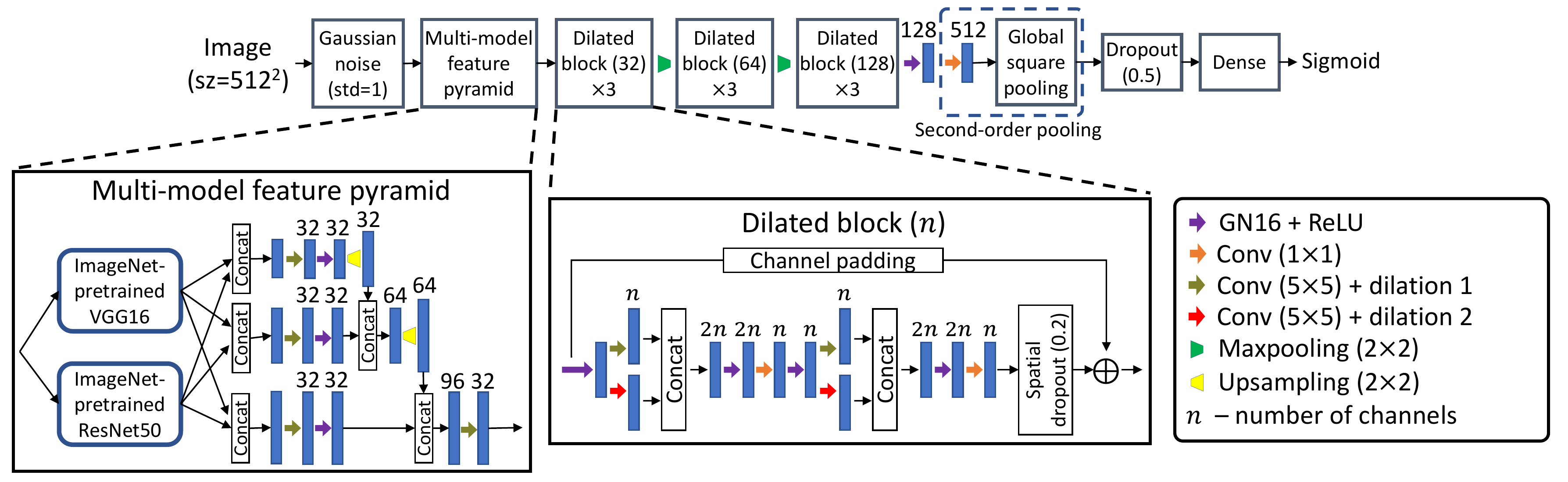}
\end{center}
   \caption{Illustration of the custom deep learning network developed for large number of label recognition problem.}
\label{network}
\end{figure}

We propose dilated blocks to learn the high-level features from the extracted ImageNet features. Each dilated block is composed of dilated convolutions for multi-scale features \cite{Journal:Yu:arXiv2015}, a skip connection of identity mapping to improve convergence \cite{Conference:He:ECCV2016}, and spatial dropout to reduce overfitting. Group normalization (16 groups) \cite{Conference:Wu:ECCV2018} whose performance is independent of the training batch size is used with ReLU. Dilated blocks with different feature channels are cascaded with maxpooling to learning more abstract features. Instead of global average pooling, second-order pooling is used, which is proven to be effective for fine-grained classification \cite{Conference:Yu:ECCV2018}. Second-order pooling maps the features to a higher-dimensional space where they can be more separable. Following \cite{Conference:Yu:ECCV2018}, the second-order pooling is implemented as a 1$\times$1 convolution followed by global square pooling (Figure~\ref{network}).

Image augmentation with rigid transformations is used to avoid overfitting. As most of an image should be included, we limit the augmentation to rotation ($\pm$\ang{10}) and shifting ($\pm${10}\%). The probability of an image to be transformed is 80\%. The optimizer Nadam is used with a learning rate of 2$\times$10$^{-6}$, a batch size of 48, and 20 epochs. To ensure efficient learning, we developed two instances of this network, one for the core finding labels (CFL labels) and the other for detailed FFL labels that have a support of at least 100 images for training to exploit the mutually reinforcing nature of the coarse-fine labels. 
Due to the variability in the size of the dataset per FFL label, the AUC per FFL label is not always a good indicator for precision on a per image level as it is dominated by the negative examples. To ensure we report as few irrelevant findings while still detecting all critical findings within an image, we select operating points on the ROC curves per label based on optimizing the F1 score, a well-known measure of accuracy, as
\begin{equation}
L(\theta)=-ln(\frac{1}{n}\sum_{i=1}^{n}F1_{i}(\theta))
\end{equation}

\subsection{FFL Pattern-Report Database Creation}
Using the FFL label detection algorithm, we can describe a report (its relevant sections) as a binary pattern vector $P=\{I_{P}(F_{j})\}$ where $I_{P}(F_{j})=1$ if the FFL label $F_{j}\in \hat{F}$ is present in the report and zero otherwise. Here $\hat{F}$ is the set of FFL labels used in training the deep learning models. During the database creation process, we collect all reports characterized by the same binary pattern vector, and rank them based on the support provided by their constituent sentences. Let $\hat{R_{P}}={r_{s}}$ be the collection of reports spanned by a pattern vector $P$. Then
\begin{equation}
Rank(r_{s})=\sum_{j=1}^{M_{s}}h(s_{j})
\label{rank}
\end{equation}
where $M_{s}$ is the number of relevant sentences in report $r_{s}$ spanned by one or more of the FFL labels in the pattern $P$. Here $h(s_{j})$ is given by
$h(s_{j})=\frac{\mbox{Number of reports }r_{i}\mbox{ that contain }s_{j}}{|\hat{R_{P}}|}$
The highest ranked reports are then stored as associated reports with the binary pattern vectors in a database. 

\subsection{Report assembly}
\begin{figure}
\begin{center}
   \includegraphics[width=0.8\linewidth]{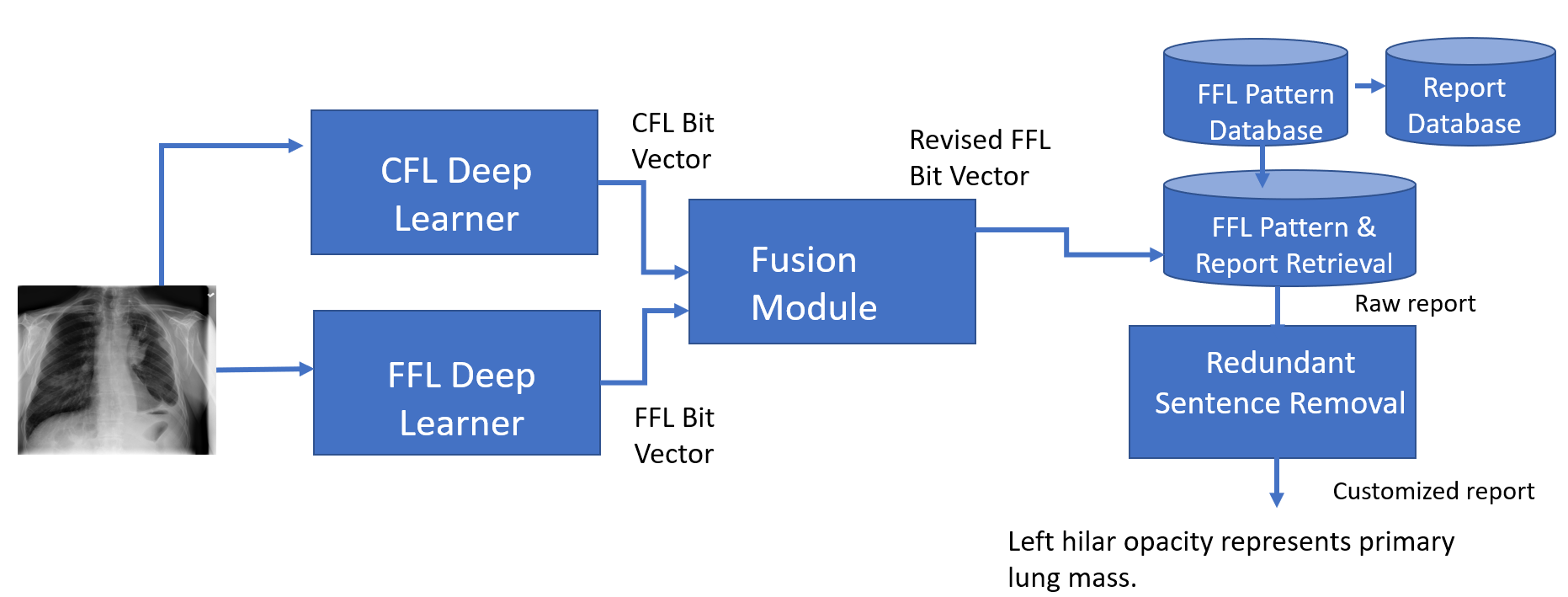}
\end{center}
   \caption{Illustration of the report generation algorithm.}
\label{overall}
\end{figure}

The overall report generation workflow is illustrated in Figure~\ref{overall}. An image is fed to the two deep learning networks built for CFL and FFL patterns and their predictions thresholded using the image-based precision-recall F1-score for optimization. The resulting pattern vectors are combined to result in the consolidated FFL pattern vector $Q=\{I_{Q}(F_{j})\}$. The best matching reports are then derived from the semantically nearest pattern vectors in the database. The semantic distance between the query FFL bit pattern vector $Q$ and a matching pattern vector from the database $P$ is given by 
\begin{equation}
d(Q,P)=\frac{\sqrt{\sum_{l=1}^{|\hat{F}|}w_{l}(I_{P}(F_{l})-I_{Q}(F_{l}))^{2}}}{|\hat{F}|}
\label{nearest}
\end{equation}

\begin{figure}[h]
\begin{center}
   \includegraphics[width=0.7\linewidth]{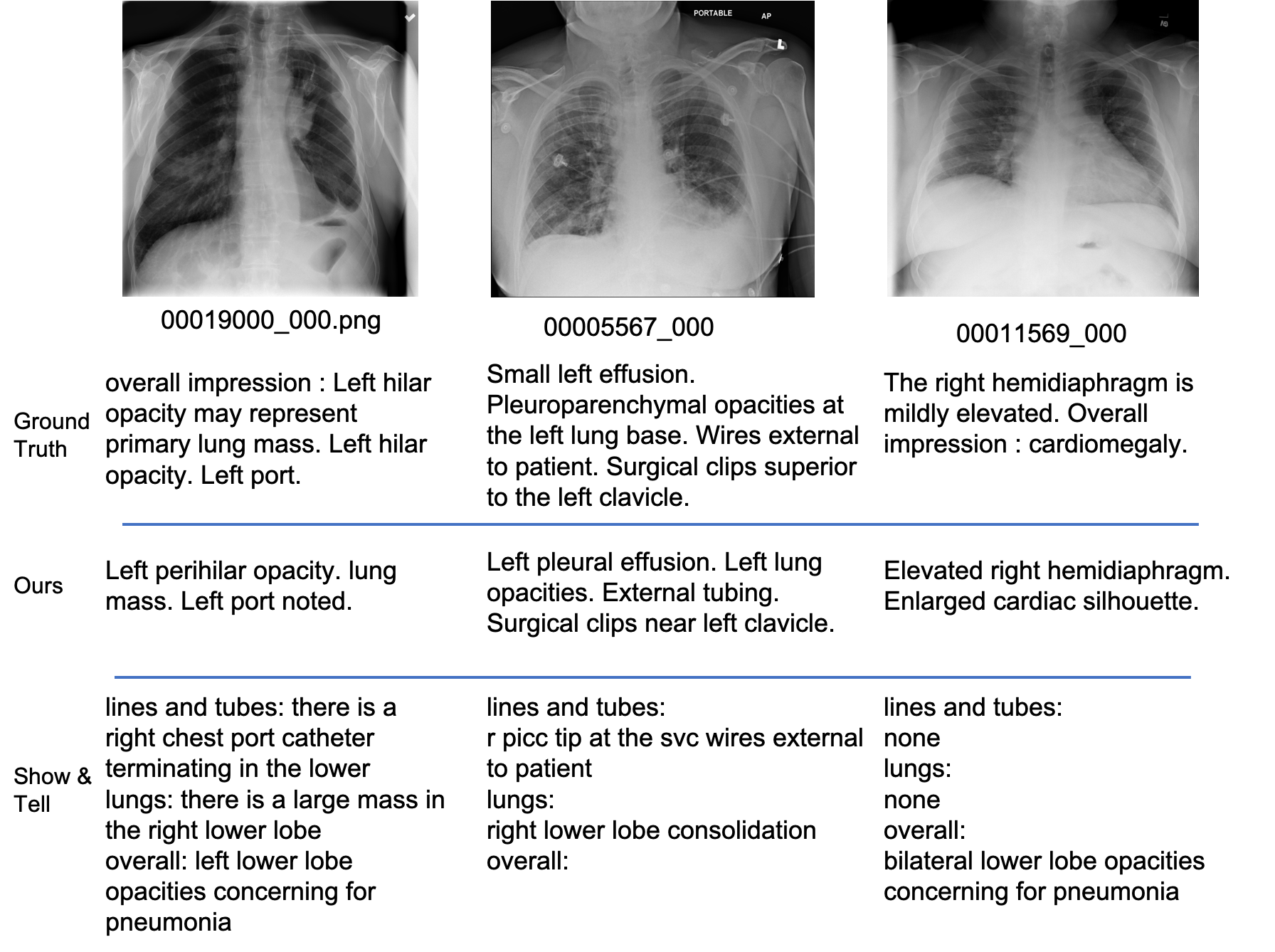}
\end{center}
   \caption{Illustration of quality of reports generated by different methods.}
\label{resultexample}
\end{figure}
where $w_{l}$ is the weight associated with the FFL label $F_{l}$.  A criticality rank for each core findings on a scale of 1 to 10 was supplied by the clinicians which was normalized and used to weigh the clinical importance of a finding during matching. Once the matching FFL pattern is determined, the highest ranked report as given by Equation~\ref{rank} associated with the FFL pattern is retrieved as the best matching report. Finally, we drop all sentences from the retrieved report whose evidence cannot be found in the FFL label pattern of the query thus achieving the variety needed in returned reports per query. Although with 457 FFL labels, the number of possible binary patterns would be large ($2^{457}$), due to the sparseness of 5-7 findings per report, the actual number of distinct binary patterns in the database of over 232,000 reports was only 924 patterns corresponding to 5246 distinct sentences in the precomputed ranked list across all patterns. Thus the lookup per predicted pattern is a fairly trivial operation which is O(1) with indexing and takes less than 5 msec.

\section{Results}
We collected datasets from three separate sources, namely, the MIMIC \cite{mimic-4} dataset of over 220,000 reports with associated frontal images, 2964 unique Indiana reports \cite{indiana} with associated images and a set of 10,000 NIH \cite{wang2017chestx} released images re-read by our team of radiologists to produce a total of 232,964 image-report pairs for our experiments.

\begin{table}
\begin{center}
\begin{tabular}{|c|c|c|c|c|c|c|}
\hline
Dataset	& FFL Label Phases&	Train&	Validate&	Test & Average  AUC &  Weighted  \\
\hline
MIMIC-4 + NIH	&CFL labels&	249,286&35,822&	70,932&0.805  & 0.841\\
\hline
MIMIC-4+NIH	&FFL labels&	75,613	&10,615&	20,941 &0.729  & 0.716 \\
\hline
\end{tabular}
\caption{Illustration of the datasets and performance of fine grained classification model for CFL and FFL labels (last column is average of AUCs weighted by the number of samples per each category). }
\label{datasets}
\end{center}
\end{table}

\textbf{Evaluating FFL label extraction accuracy: }We evaluated the accuracy of FFL label extraction by noting the number of findings missed and overcalled (which included errors in negation sense detection) as well the correctness and completeness of association of modifiers with the relevant core findings. The result of evaluation for the Indiana dataset \cite{indiana} by our clinicians is shown in Table~\ref{fllevaluation}. As can be seen, the FFL label extraction is highly accurate in terms of the coverage of findings with around 3\% error mostly due to negation sense detection. Further, the association of modifiers to core findings given by the phrasal grouping algorithm is also accurate with over 99\% precision and recall.   

\textbf{FFL Label Prediction from Deep Learning: }The training, validation and test image datasets used for building the CFL and FFL models used MIMIC-4 and NIH datasets as shown in Table~\ref{datasets}. The AUC averaged for all CFL labels and FFL labels is also shown in that table. In addition, using the F1-score-based optimization, the mean average image-based precision for CFL labels was 73.67\% while the recall was 70.6\%.


\begin{table}
\begin{center}
\begin{scriptsize}
\begin{tabular}{|c|c|c|c|c|c|c|}
\hline
reports & relevant  & FFL patterns  & missed  & overcall & incorrect association  & missed\\
analyzed & sentences & extracted & findings & (negated findings) & of modifiers  & modifiers \\
\hline
2964             & 3046               & 5245                   & 0               & 168                         & 49                                 & 11  
\\
\hline
\end{tabular}
\end{scriptsize}
\end{center}
\caption{The accuracy of FFL label extraction from reports.}
\label{fllevaluation}
\end{table}

\textbf{Evaluation of report generation: } Due to the ontological mapping used to abstract the description of findings, the match produced from our approach is at a more semantic level rather than lexical in comparison to other approaches.  Figure~\ref{resultexample} shows the reports manually and automatically produced by our approach and a comparative approach implemented from a visual attention-based captioning model \cite{vinyals2015show}. 
We compared the performance of our algorithm with several state-of-the-art baselines from recent literature \cite{vinyals2015show,multimodalrecurrent,li2019knowledge,szolovits,jing2017automatic,jiebo}. These included a range of approaches from visual attention-based captioning \cite{vinyals2015show}, knowledge-driven report generation \cite{li2019knowledge}, clinically accurate report generation \cite{szolovits}, to a strawman approach using a set of template sentences manually chosen by clinicians for the FFL labels instead of the nearest report selection algorithm described earlier. Although we have tested our algorithm for very large number of images from the combined MIMIC-NIH data, for purposes of comparison, we show the results on the same Indiana test dataset that has been used most commonly by other algorithms as reported in \cite{jing2017automatic}. The resulting performance using the popular scoring metrics is shown in Table~\ref{fflresults} showing that our algorithm outperforms other approaches in all the established scoring metrics.

\textbf{Conclusions: }We presented an explainable AI approach to semantically correct radiology report generation. The results show superior performance both because of the detailed descriptive nature of labels, and due to a statistically informed report retrieval process that ensures a semantic match.

\begin{table}
\begin{center}
\begin{tabular}{|c|c|c|c|c|c|c|}
\hline
Methods	&BLEU-1	&BLEU-2	&BLEU-3	&BLEU-4	&METEOR	&ROUGE-L\\
\hline
Vis-Att \cite{vinyals2015show}	&0.39	&0.25	&0.16	&0.11	&0.16	&0.32\\
\hline
MM-Att \cite{multimodalrecurrent}	&0.46	&0.35	&0.27	&0.19	&0.27	&0.36\\
\hline
KERP \cite{li2019knowledge}	&0.48	&0.32	&0.22	&0.16	&-	&0.33\\
\hline
Template-based	&0.28	&0.29	&0.32	&0.27	&0.35	&0.34\\
\hline
Clinical Accurate \cite{szolovits}	&0.35	&0.22	&0.15	&0.1	&-	&0.45\\
\hline
Co-Att \cite{jing2017automatic}	&0.51	&0.39	&0.30	&0.25	&0.21	&0.44\\
\hline
Jiebo Luo \cite{jiebo}	&0.53	&0.37	&0.31	&0.25	&0.34	&0.45\\
\hline
CFL-only 	& 0.49	& 0.39	& 0.36	&0.32 &0.48	&0.52  \\
\hline
FFL+CFL -based (ours)	&\bf{0.56}	&\bf{0.51}	&\bf{0.5	}&\bf{0.49}	&\bf{0.55}	&\bf{0.58}\\
\hline
\end{tabular}
\caption{Comparative performance of report generation by various methods.}
\label{fflresults}
\end{center}
\end{table}

{\small
\bibliographystyle{splncs03}
\bibliography{paper2018}

\begin{thebibliography}{10}
\providecommand{\url}[1]{\texttt{#1}}
\providecommand{\urlprefix}{URL }

\bibitem{jamasubmission}
Blocked, A.: Artificial intelligence versus entry-level radiologists for
  full-fledged preliminary read of frontal ap chest radiographs: A comparative
  study. JAMA  (2020)

\bibitem{dla}
Coden, A., Gruhl, D., Lewis, N., Tanenblatt, M., Terdiman, J.: Spot the drug!
  an unsupervised pattern matching method to extract drug names from very large
  clinical corpora. In: Proceedings of the 2012 IEEE Second International
  Conference on Healthcare Informatics, Imaging and Systems Biology. pp.
  7008--7024 (2012)

\bibitem{indiana}
Demmer-Fushma, D., et~al.: Preparing a collection of radiology examinations for
  distribution and retrieval. Journal of American Medical Informatics
  Association (JAMIA)  23(2),  304--310 (2014)

\bibitem{gale2018producing}
Gale, W., Oakden-Rayner, L., Carneiro, G., Bradley, A.P., Palmer, L.J.:
  Producing radiologist-quality reports for interpretable artificial
  intelligence. arXiv preprint arXiv:1806.00340  (2018)

\bibitem{guo2017efficient}
Guo, Y., Kakrania, D., Baldwin, T., Syeda-Mahmood, T.: Efficient clinical
  concept extraction in electronic medical records. In: Thirty-First AAAI
  Conference on Artificial Intelligence (2017)

\bibitem{hansell2008fleischner}
Hansell, D.M., Bankier, A.A., MacMahon, H., McLoud, T.C., Muller, N.L., Remy,
  J.: Fleischner society: glossary of terms for thoracic imaging. Radiology
  246(3),  697--722 (2008)

\bibitem{harzig2019addressing}
Harzig, P., Chen, Y.Y., Chen, F., Lienhart, R.: Addressing data bias problems
  for chest x-ray image report generation. arXiv preprint arXiv:1908.02123
  (2019)

\bibitem{Conference:He:CVPR2016}
He, K., Zhang, X., Ren, S., Sun, J.: Deep residual learning for image
  recognition. In: IEEE Conference on Computer Vision and Pattern Recognition.
  pp. 770--778 (2016)

\bibitem{Conference:He:ECCV2016}
He, K., Zhang, X., Ren, S., Sun, J.: Identity mappings in deep residual
  networks. In: European Conference on Computer Vision. LNCS, vol. 9908, pp.
  630--645 (2016)

\bibitem{jing2017automatic}
Jing, B., Xie, P., Xing, E.: On the automatic generation of medical imaging
  reports. arXiv preprint arXiv:1711.08195  (2017)

\bibitem{mimic-4}
Johnson, A.E.W., Pollard, T.J., Berkowitz, S.J., Greenbaum, N.R., Lungren,
  M.P., y.~Deng, C., Mark, R.G., Horng, S.: Mimic-cxr: A large publicly
  available database of labeled chest radiographs. arXiv preprint
  arXiv:1901.07042  (2019)

\bibitem{krause2017hierarchical}
Krause, J., Johnson, J., Krishna, R., Fei-Fei, L.: A hierarchical approach for
  generating descriptive image paragraphs. In: Proceedings of the IEEE
  Conference on Computer Vision and Pattern Recognition. pp. 317--325 (2017)

\bibitem{li2019knowledge}
Li, C.Y., Liang, X., Hu, Z., Xing, E.P.: Knowledge-driven encode, retrieve,
  paraphrase for medical image report generation. arXiv preprint
  arXiv:1903.10122  (2019)

\bibitem{li2018hybrid}
Li, Y., Liang, X., Hu, Z., Xing, E.P.: Hybrid retrieval-generation reinforced
  agent for medical image report generation. In: Advances in Neural Information
  Processing Systems. pp. 1530--1540 (2018)

\bibitem{Conference:Lin:CVPR2017}
Lin, T.Y., Doll{\'a}r, P., Girshick, R., He, K., Hariharan, B., Belongie, S.:
  Feature pyramid networks for object detection. In: IEEE Conference on
  Computer Vision and Pattern Recognition. pp. 2117--2125 (2017)

\bibitem{szolovits}
Liu, G., et~al.: Clinically accurate chest x-ray report generation.
  arXiv:1904.02633v  (2019)

\bibitem{esg}
M.C.McCord, J.W.Murdock, B.K.Bogurae: Deep parsing in watson. IBM. Jl. of
  Research and Development  56(3),  5--15 (2012)

\bibitem{Conference:Nguyen:ISCAS2018}
Nguyen, L.D., Lin, D., Lin, Z., Cao, J.: Deep {CNNs} for microscopic image
  classification by exploiting transfer learning and feature concatenation. In:
  IEEE International Symposium on Circuits and Systems. pp. 1--5 (2018)

\bibitem{rennie2017self}
Rennie, S.J., Marcheret, E., Mroueh, Y., Ross, J., Goel, V.: Self-critical
  sequence training for image captioning. In: Proceedings of the IEEE
  Conference on Computer Vision and Pattern Recognition. pp. 7008--7024 (2017)

\bibitem{Journal:Simonyan:arXiv2014}
Simonyan, K., Zisserman, A.: Very deep convolutional networks for large-scale
  image recognition. arXiv:1409.1556 [cs.CV]  (2014)

\bibitem{vinyals2015show}
Vinyals, O., Toshev, A., Bengio, S., Erhan, D.: Show and tell: A neural image
  caption generator. In: Proceedings of the IEEE conference on computer vision
  and pattern recognition. pp. 3156--3164 (2015)

\bibitem{wang2017chestx}
Wang, X., Peng, Y., Lu, L., Lu, Z., Bagheri, M., Summers, R.M.: Chestx-ray8:
  Hospital-scale chest x-ray database and benchmarks on weakly-supervised
  classification and localization of common thorax diseases. In: Proceedings of
  the IEEE conference on computer vision and pattern recognition. pp.
  2097--2106 (2017)

\bibitem{wang2018tienet}
Wang, X., Peng, Y., Lu, L., Lu, Z., Summers, R.M.: Tienet: Text-image embedding
  network for common thorax disease classification and reporting in chest
  x-rays. In: Proceedings of the IEEE conference on computer vision and pattern
  recognition. pp. 9049--9058 (2018)

\bibitem{Conference:Wu:ECCV2018}
Wu, Y., He, K.: Group normalization. In: European Conference on Computer
  Vision. pp. 3--19 (2018)

\bibitem{xu2015show}
Xu, K., Ba, J., Kiros, R., Cho, K., Courville, A., Salakhudinov, R., Zemel, R.,
  Bengio, Y.: Show, attend and tell: Neural image caption generation with
  visual attention. In: International conference on machine learning. pp.
  2048--2057 (2015)

\bibitem{multimodalrecurrent}
Xue, Y., et~al.: Multimodal recurrent model with attention for automated
  radiology report generation. In: International Conference on Medical Image
  Computing and Computer-Assisted Intervention. LNCS, vol. 11072, pp. 457--466
  (2018)

\bibitem{Journal:Yu:arXiv2015}
Yu, F., Koltun, V.: Multi-scale context aggregation by dilated convolutions.
  arXiv:1511.07122 [cs.CV]  (2015)

\bibitem{Conference:Yu:ECCV2018}
Yu, K., Salzmann, M.: Statistically-motivated second-order pooling. In:
  European Conference on Computer Vision. pp. 600--616 (2018)

\bibitem{jiebo}
Yua, J., Liao, H., Luo, R., Luo, J.: Automatic radiology report generation
  based on multi-view image fusion and medical concept enrichment. In:
  Proceedings of the International Conference on Medical Image Computing and
  Computer-Assisted Intervention. LNCS, vol. 9356, pp. 234--241 (2019)

\end{thebibliography}
}
\end{document}